\documentclass{article} % For LaTeX2e
\usepackage{iclr2027_conference,times}

\usepackage{amsmath,amsfonts,bm}

\def\eqref#1{equation~\ref{#1}}
\def\1{\bm{1}}

\DeclareMathAlphabet{\mathsfit}{\encodingdefault}{\sfdefault}{m}{sl}
\SetMathAlphabet{\mathsfit}{bold}{\encodingdefault}{\sfdefault}{bx}{n}

\usepackage{url}
\usepackage{xurl}
\usepackage{hyperref}
\usepackage{amsmath}
\usepackage{amssymb}
\usepackage{graphicx}
\usepackage{booktabs}
\usepackage{array}
\usepackage[table]{xcolor}
\usepackage{multirow}
\usepackage{wrapfig}
\usepackage{caption}

\title{The Model Knows Another Way: Strategy Switching for Effective RLVR Exploration}

\author{
    Jin Cui\textsuperscript{\rm 1},
    Xinyue Long\textsuperscript{\rm 1,\rm 2},
    Boran Zhao\textsuperscript{\rm 1,\rm 2}\thanks{Corresponding author},
    Pengju Ren\textsuperscript{\rm 1},
    Hao Dong\textsuperscript{\rm 3} \\
    \textsuperscript{\rm 1}State Key Laboratory of Human-Machine Hybrid Augmented Intelligence,\\
    and Institute of Artificial Intelligence and Robotics, Xi'an Jiaotong University\\
    \textsuperscript{\rm 2}School of Software Engineering, Xi'an Jiaotong University \\
    \textsuperscript{\rm 3}ELLIS Institute Finland and Tampere University  
}

\iclrfinalcopy % Uncomment for camera-ready version, but NOT for submission.
\begin{document}

\maketitle
\lhead{}

\begin{abstract}
Reinforcement learning with verifiable rewards (RLVR) is often limited by insufficient exploration: difficult problems can yield uniformly incorrect rollout groups and therefore little learning signal. We show that such failures need not reflect missing capability. Instead, finite sampling often concentrates on a problem-specific dominant reasoning strategy while leaving alternative strategies already supported by the model unexplored. Moreover, the accessibility of these strategies evolves during RL: some are internalized into autonomous behavior, while others become difficult to elicit before being absorbed. Motivated by these observations, we introduce \textbf{Problem--Strategy Rollout Allocation (PSRA)}, which treats unguided and strategy-conditioned prompts as competing exploration arms and uses Bayesian sequential allocation to direct a fixed rollout budget toward arms most likely to yield informative, non-saturated groups. A preservation objective keeps useful strategy-conditioned routes accessible while successful guided behaviors are transferred to the unguided policy. Across Qwen2.5 models from 1.5B to 7B and two RL training corpora, PSRA consistently improves reasoning performance, reduces dead saturation, strengthens out-of-distribution transfer, and maintains larger gains under increased inference budgets.

\end{abstract}

\section{Introduction}
\label{sec:introduction}

Reinforcement learning with verifiable rewards (RLVR) has become a central paradigm for improving the reasoning performance of large language models. Yet its gains remain strongly constrained by exploration. Growing evidence suggests that RL often sharpens reasoning trajectories already accessible to the base model: pass@1 can improve substantially while pass@k or broader solution coverage improve much less, while concentrating probability mass onto a narrower set of reasoning modes \citep{yue2025does,wu2025invisible,lee2026sage}. This limitation is particularly consequential on hard problems, where uniformly unsuccessful rollout groups provide little or no learning signal. Recent approaches therefore introduce solution prefixes, hints, privileged contexts, or strategy-level conditions to make successful trajectories reachable \citep{yan2025luffy,nath2025adaptive,zhang2026scaf,wang2026hint,agrawal2026offcontext,lee2026nudging}. While effective, these methods leave open a more basic question: \textbf{\emph{what should the model be guided toward when its default exploration fails?}}

We study this question at the level of \emph{reasoning strategies}: reusable, high-level solution procedures already supported by the model. Clustering internal representations of base-model rollouts reveals a compact repertoire of semantically coherent strategies, which can be reliably re-elicited through natural-language instructions. Crucially, their use is strongly problem-conditioned. For a given problem, finite autonomous sampling typically concentrates on one dominant strategy direction, leaving other members of the same repertoire weakly selected. This concentration becomes costly when the dominant route fails: on problems where the initial rollout group is entirely incorrect, allocating the same additional budget to another strategy recovers successful trajectories substantially more often than either continuing along the dominant strategy or simply sampling more from the default policy. Thus, an all-failure group need not indicate the absence of a viable solution procedure; it can instead reflect a finite-budget \emph{strategy-selection failure}. This is consistent with evidence that RL primarily selects among reasoning patterns acquired beforehand \citep{krishnamurthy2026select}, and conditioning can expose modes neglected by naive sampling \citep{lee2026nudging,xu2026imax}.

Strategy-conditioned exploration, however, introduces a second challenge: the available exploration directions themselves evolve during RL. Tracking fixed strategy conditions across checkpoints reveals two distinct dynamics. Some strategies become increasingly expressed without guidance, indicating successful internalization, whereas others lose their ability to redirect the policy without a corresponding increase in autonomous use. A strategy is therefore not merely a one-time sampling scaffold, but a control handle whose usefulness must persist long enough to be exploited and transferred. This distinction matters for existing guided-RL methods. Without explicitly maintaining the conditional distribution, useful alternative routes may disappear before their behaviors are absorbed. Effective strategy-guided exploration must therefore address both \emph{where to explore} and \emph{how to keep those exploration directions available}.

Motivated by these observations, we introduce \textbf{\textit{Problem--Strategy Rollout Allocation (PSRA)}}, which treats the unguided prompt and each strategy-conditioned prompt as competing exploration arms. For each problem--strategy pair $(q,z)$, PSRA first collects a small probe and constructs a hierarchical Bayesian posterior over its success probability, incorporating problem difficulty, global strategy effectiveness, and the predicted reasoning mass left uncovered by finite sampling. It then allocates the remaining fixed rollout budget to maximize the expected number of non-saturated groups, directing computation toward arms most likely to provide informative group-relative gradients. Because the unguided condition participates in the same objective, strategy guidance is selected only when it offers greater expected learning value than continued autonomous sampling, avoiding hand-designed failure triggers or fixed routing rules. Successful guided trajectories are transferred to the unguided policy through importance-corrected updates, while an on-context preservation term maintains the strategy-conditioned routes that made those trajectories reachable. 

Experiments across models with varying scales and two RL training corpora consistently validate this design. PSRA recovers $26\%$ of initially unsolved problems versus $17\%$ for GRPO, reduces dead saturation while preserving higher policy entropy, and its advantage generally increases with larger pass@$k$ budgets, indicating broader strategy coverage. Our contributions are summarized as:
\begin{itemize}
    \item We provide a strategy-level account of RLVR exploration that LLMs contain a steerable repertoire of reasoning strategies, yet finite autonomous sampling concentrates strongly on problem-specific dominant routes, while the accessibility of alternative routes can either be internalized or selectively lost during RL.
    
    \item We introduce \textbf{PSRA}, a Bayesian allocation framework over problem--strategy arms that targets non-saturated learning signals while jointly transferring successful guided behaviors to the unguided policy and preserving useful strategy-conditioned exploration directions.
    
    \item Experiments across model scales and training distributions, showing consistent improvements over vanilla GRPO and guided-RL baselines, stronger recovery of initially unsolved problems, reduced saturation, and increasing advantages under larger inference budgets.
\end{itemize}

\vspace{-2mm}

\section{Related Work}

\paragraph{Learning signals in RLVR.}
GRPO relies on within-group reward variation, so all-correct or all-incorrect groups yield zero relative advantages and provide no reward-driven update \citep{Shao2024DeepSeekMathPT}.
DAPO alleviates this degeneracy through dynamic sampling that filters saturated groups and resamples prompts to maintain informative batches \citep{Yu2025DAPOAO}.
Such methods improve the availability of learning signals, but an all-failure group is typically treated as an uninformative prompt rather than a recoverable exploration failure.
PSRA instead leverage another reasoning strategy already available to the model to turn the problem into a non-saturated, learnable group.

\paragraph{Guided RLVR exploration.}
A growing line of work uses external guidance to recover successful reasoning trajectories when on-policy exploration fails.
LUFFY mixes off-policy expert traces with on-policy rollouts \citep{yan2025luffy}; Guide adaptively introduces guidance on difficult problems \citep{nath2025adaptive}; BREAD branches from partial expert anchors \citep{Zhang2025BREADBR}; and HINT uses abstract Meta-Hints with affinity-aware optimization to reduce mismatch between guidance and the policy \citep{wang2026hint}.
NudgeRL conditions rollouts on strategy-level contexts \citep{lee2026nudging}, while Uniqueness-Aware RL upweights correct trajectories that instantiate rare observed strategies \citep{Hu2026RewardingTR}.
These methods establish the value of guided or diverse reasoning, but largely treat the guidance as given or act on patterns already sampled. PSRA instead constructs a reusable strategy repertoire from the model itself, estimates its problem-specific exploration value online, and explicitly preserves useful strategy conditions as the policy evolves.

\vspace{-2mm}

\section{From Strategy to Problem-Conditioned Exploration}
\label{sec:strategy_analysis}

\begin{figure}
    \centering
    \includegraphics[width=1\linewidth]{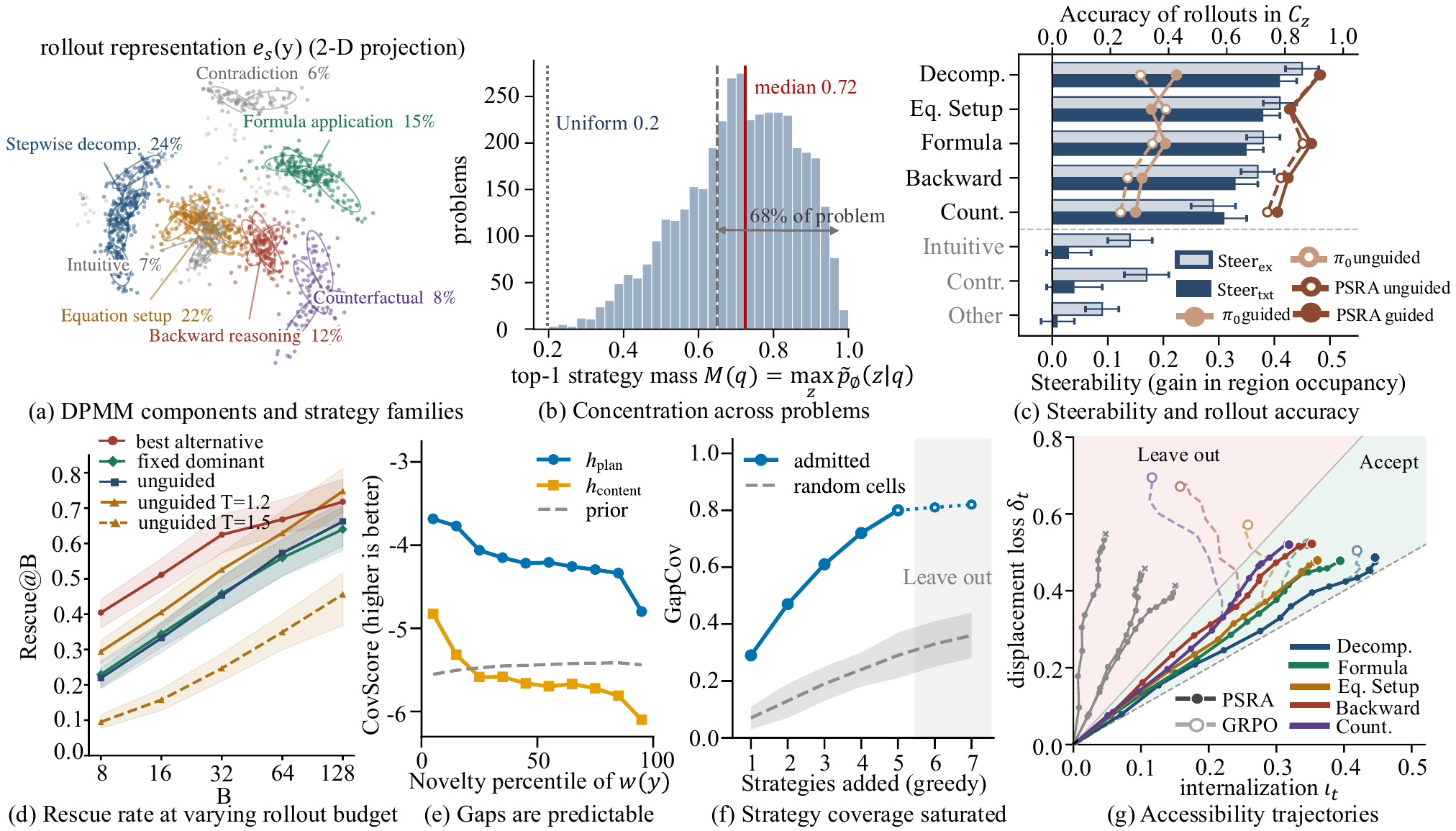}
    \caption{\textbf{Diagnosing strategy-level exploration in the base policy.}
    (a) Representation-space clustering reveals a compact set of reasoning strategies.
    (b) Autonomous rollouts are strongly concentrated.
    (c) Strategy conditioning reliably redirects generation toward distinct reasoning regions.
    (d) Switching to the best alternative strategy improves rescue rate over continued dominant-strategy or unguided sampling.
    (e) Pre-generation states predict reasoning directions missed by finite rollouts.
    (f) A small set of certified strategies covers most of the predicted exploration gaps.
    (g) During RL, strategy accessibility exhibits both internalization and displacement.}
    \label{fig:analysis}
    \vspace{-4mm}
\end{figure}

Pretraining equips the base policy with a compact repertoire of stable reasoning strategies. We first identify these strategies, then examine how finite-budget autonomous sampling selects among them and where alternative strategies provide useful exploration opportunities.

\paragraph{Reusable pretrained strategy repertoire.}
We re-encode base-model rollouts under their original problem context and pool intermediate hidden states into rollout representations $e_{\mathrm{s}}(y)$. A Dirichlet process mixture model (DPMM) partitions these representations into fine-grained components, which are merged into semantic strategy candidates according to their solution structure (Fig.~\ref{fig:analysis}a).
We behaviorally certify each candidate using both exemplar and natural-language instruction conditioning, retaining only strategies that consistently redirect rollouts toward their associated representation regions. Five of seven candidates satisfy this criterion and form the strategy set $\mathcal Z$; natural-language instructions retain 88\% of the displacement induced by full exemplars, while matched placebo instructions have negligible effect (Fig.~\ref{fig:analysis}c).

For each strategy $z$, let $\mathcal K_z$ denote its associated DPMM components. We estimate intervention lift
\begin{equation}
\ell_z(k)
=
\frac{P(k(y)=k\mid z)}
     {P(k(y)=k\mid\varnothing)},
\end{equation}
and define its footprint $C_z\subseteq\mathcal K_z$ by accumulating the highest-lift components until they account for 50\% of the strategy-associated guided mass.

Repeating the same analysis across Qwen2.5-1.5B, 3B, and 7B recovers the same strategy families across scales, with only slightly weaker elicitation stability on the 1.5B model, providing preliminary evidence that the discovered repertoire captures a shared and sufficiently complete strategy basis.

\paragraph{Problem-conditioned strategy selection.}
For a problem $q$, define the autonomous occupancy of strategy $z$ as
\begin{equation}
p_{\varnothing}(z\mid q)
=
P\!\left(k(y)\in C_z\mid q,\varnothing\right),
\qquad
z_{\mathrm{dom}}(q)
=
\arg\max_{z\in\mathcal Z}
\frac{p_{\varnothing}(z\mid q)}
{\sum_{z'\in\mathcal Z}p_{\varnothing}(z'\mid q)} .
\end{equation}
Autonomous generation is strongly concentrated at the problem level: most strategy-associated mass is typically assigned to a single dominant strategy (Fig.~\ref{fig:analysis}b), leaving alternative directions weakly explored. This becomes consequential when the dominant route fails. On problems for which the initial rollout group contains no correct solution, allocating the same additional budget to the dominant strategy or to unguided sampling recovers substantially fewer problems than switching to the best alternative strategy (Fig.~\ref{fig:analysis}d).
Thus, a zero-reward group can reflect a finite-budget strategy-selection failure rather than the absence of a viable reasoning procedure.

\paragraph{Predicting missed strategy opportunities.}
The usefulness of a strategy depends on both the problem and what finite sampling has already covered. We therefore train a lightweight predictor $G$ from the pre-generation planning state $h_{\mathrm{plan}}(q)$ to a distribution $\hat q(k\mid q)$ over frozen DPMM components. On held-out problems, we evaluate $G$ with a novelty-weighted log score that emphasizes trajectories far from the observed rollout group; the planning representation substantially outperforms an early-layer content representation, including on the most novel trajectories (Fig.~\ref{fig:analysis}e). Refer to Appendix \ref{app:strategy_analysis} for more details and training procedure.

We define the high-probability components left unvisited by an observed rollout set $O$ as
\begin{equation}
U_m(q;O)
=
\operatorname{Top}_m
\left\{
\hat q(k\mid q):
k\notin\mathrm{hit}(O)
\right\},
\end{equation}
and the remaining opportunity associated with strategy $z$ as
\begin{equation}
\Omega(q,z;O)
=
\sum_{k\in U_m(q;O)\cap C_z}
\hat q(k\mid q).
\end{equation}
The five certified strategies account for most of the predicted gap mass, far exceeding matched random component sets, with coverage saturating rapidly as strategies are added (Fig.~\ref{fig:analysis}f). Hence, a small strategy basis captures most structured exploration opportunities exposed by finite sampling.

\paragraph{Strategy accessibility evolves during RL.}
Strategy-guided exploration additionally requires that a strategy remain an effective control handle as the policy changes. With the initial footprints $C_z^{(0)}$ frozen, we measure strategy-induced displacement as
\begin{equation}
D_t(z)
=
\mathrm{JS}
\left(
P_t(k\mid z)
\,\Vert\,
P_t(k\mid\varnothing)
\right),
\end{equation}
and decompose its reduction into
\begin{equation}
\delta_t(z)
=
1-\frac{D_t(z)}{D_0(z)},
\qquad
\iota_t(z)
=
\frac{
P_t(C_z^{(0)}\mid\varnothing)
-
P_0(C_z^{(0)}\mid\varnothing)
}{
P_0(C_z^{(0)}\mid z)
-
P_0(C_z^{(0)}\mid\varnothing)
},
\qquad
L_t(z)=\delta_t(z)-\iota_t(z).
\end{equation}
Here $\iota_t$ measures internalization into autonomous behavior, while $L_t$ captures residual loss of conditional accessibility.
Both dynamics occur during standard RLVR: some strategies are progressively internalized, whereas others lose steerability before their behaviors are absorbed (Fig.~\ref{fig:analysis}g).

Together, these observations identify the problem--strategy pair $(q,z)$ as the natural unit of structured exploration: $\Omega(q,z;O)$ captures the remaining coverage opportunity, $p_t(q,z)$ its current utility, and $L_t(z)$ whether the corresponding strategy remains accessible.

\section{Problem--Strategy Rollout Allocation}
\label{sec:method}

\begin{figure}
    \centering
    \includegraphics[width=1\linewidth]{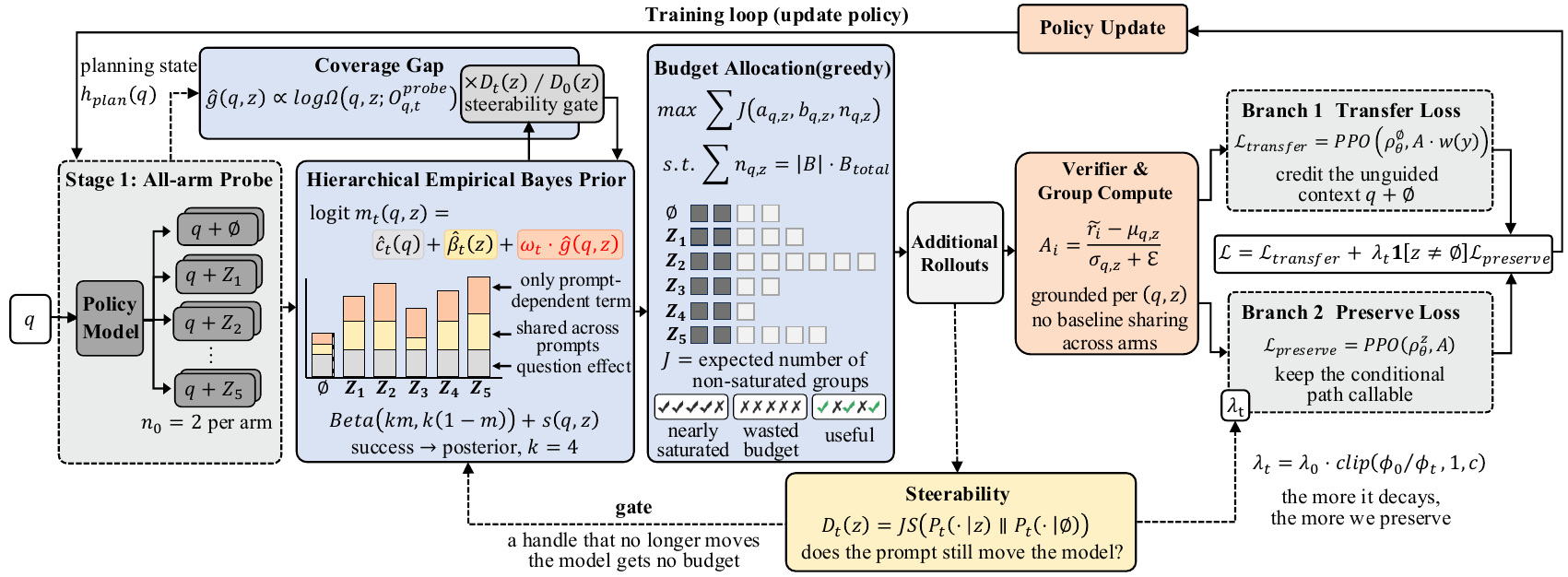}
    \caption{\textbf{Overview of PSRA Structure.}
    PSRA allocates rollout budget over unguided and strategy-conditioned arms, then transfers successful guided behaviors while preserving strategy accessibility.}
    \label{fig:main}
    \vspace{-4mm}
\end{figure}

Section~\ref{sec:strategy_analysis} provides a fixed strategy set $\mathcal Z$ and its representation-space footprints.
PSRA treats these strategies together with the unguided condition as exploration arms,
\begin{equation}
\mathcal A=\{\varnothing\}\cup\mathcal Z,
\qquad
p_t(q,z)
=
\Pr_{y\sim\pi_{\theta_t}(\cdot\mid c(q,z))}
[r(q,y)=1],
\end{equation}
where $c(q,z)$ is the context for problem $q$ under arm $z$.
The goal is to allocate a fixed rollout budget across $(q,z)$ pairs whose success probabilities are unknown and evolve during training.

\subsection{Bayesian Allocation over Problem--Strategy Arms}
\label{sec:allocation}

% \paragraph{Probe and posterior.}
% At each step, PSRA draws $n_0$ rollouts from every active $(q,z)$ arm and observes $s_{q,z}$ successes.
% We use an empirical-Bayes prior whose mean combines current problem difficulty, global strategy effectiveness, and the remaining strategy-addressable opportunity:
% \vspace{-1mm}
% \begin{equation}
% m_t(q,z)
% =
% \sigma\!\left(
% \hat c_t(q,z)+\hat\beta_t(z)+\omega_t\hat g_t(q,z)
% \right).
% \label{eq:prior_mean}
% \end{equation}
% Here $\hat c_t$ is a leave-one-out estimate of problem difficulty, $\hat\beta_t$ tracks the global effect of strategy $z$, and $\hat g_t$ captures its uncovered problem-specific opportunity. Let $\hat q_t(k\mid q)$ be the pre-generation component predictor from Section~\ref{sec:strategy_analysis}, and $\tilde C_z^{(t)}$ the current footprint of strategy $z$. We define

% \vspace{-2mm}
% \begin{equation}
% \begin{aligned}
% H_{-z}(q)
% &=
% \bigcup_{z'\neq z}
% \{k(y):y\in\mathrm{probe}(q,z')\},\\
% \Omega_t(q,z)
% &=
% \frac{D_t(z)}{D_0(z)}
% \sum_{k\in\tilde C_z^{(t)}\setminus H_{-z}(q)}
% \hat q_t(k\mid q),
% \qquad z\neq\varnothing ,
% \end{aligned}
% \label{eq:online_opportunity}
% \end{equation}

% and set $\hat g_t(q,z)$ to the centered log transform of $\Omega_t(q,z)$, with
% $\hat g_t(q,\varnothing)=0$.
% The factor $D_t(z)/D_0(z)$ discounts strategies whose steerability has degraded, while $\omega_t\geq0$ is fitted online and suppresses the coverage term when it is not predictive of success.

\paragraph{Probe and posterior.}
At each step, PSRA draws $n_0$ rollouts from every active $(q,z)$ arm and
observes $s_{q,z}$ successes. We use an empirical-Bayes prior whose mean combines
problem difficulty, global arm effectiveness, and the remaining strategy-addressable
opportunity:
\begin{equation}
m_t(q,z)=
\sigma\!\left(
\hat c_t^{-z}(q)+\hat\beta_t(z)+\omega_t\hat g_t(q,z)
\right).
\tag{8}
\end{equation}
Here $\hat c_t^{-z}(q)$ is a leave-one-arm-out estimate of problem difficulty,
$\hat\beta_t(z)$ is a centered probe-only EMA of the global arm effect that tracks the global effect of strategy $z$, and $\hat g_t$ captures its uncovered problem-specific opportunity.
$\omega_t\geq0$ is an online coefficient learned from preceding
probe batches; their estimators are given in Appendix~\ref{app:online_prior}. After the all-arm probe, let
\begin{equation}
O^{\mathrm{probe}}_{q,t}
=
\bigcup_{a\in\mathcal A}\mathrm{probe}_t(q,a),
\qquad
\Omega_t(q,z)
=
\frac{D_t(z)}{D_0(z)}
\Omega(q,z;O^{\mathrm{probe}}_{q,t}),
\quad z\neq\varnothing .
\tag{9}
\end{equation}
We set $\hat g_t(q,z)$ to the centered log transform of $\Omega_t(q,z)$ and
$\hat g_t(q,\varnothing)=0$. Thus, components reached by any probed arm are removed from the remaining opportunity, while $D_t(z)/D_0(z)$ discounts strategies whose steerability has degraded.

With prior strength $\kappa$, the prior and posterior are (with $n_0=2$ and $\kappa=4$)
\begin{equation}
\begin{aligned}
p_t(q,z)
&\sim
\mathrm{Beta}\!\left(
\kappa m_t(q,z),\,
\kappa[1-m_t(q,z)]
\right),\\
a_{q,z}
&=
\kappa m_t(q,z)+s_{q,z}, \quad
b_{q,z}
=
\kappa[1-m_t(q,z)]+n_0-s_{q,z}.
\end{aligned}
\label{eq:posterior}
\end{equation}

\vspace{-2mm}

\paragraph{Adaptive rollout allocation.}
Let $U_{q,z}(k)$ denote the posterior probability that arm $(q,z)$ is non-saturated after receiving $k$ additional rollouts. For binary rewards,
\begin{equation}
U_{q,z}(k)
=
\begin{cases}
1,
& 0<s_{q,z}<n_0,\\[2pt]
1-\dfrac{B(a_{q,z},\,b_{q,z}+k)}
        {B(a_{q,z},\,b_{q,z})},
& s_{q,z}=0,\\[8pt]
1-\dfrac{B(a_{q,z}+k,\,b_{q,z})}
        {B(a_{q,z},\,b_{q,z})},
& s_{q,z}=n_0,
\end{cases}
\label{eq:effective_prob}
\end{equation}
where $B(\cdot,\cdot)$ is the beta function.
Mixed probes are already informative, whereas saturated arms are valuable only if their posterior assigns sufficient probability to the opposite outcome.

PSRA allocates the remaining budget by

% \begin{equation}
% \max_{\{k_{q,z}\}}
% \sum_{q,z} U_{q,z}(k_{q,z})
% \quad
% \mathrm{s.t.}
% \quad
% \sum_{q,z} k_{q,z}
% \leq B_{\mathrm{extra}},
% \qquad
% 0\leq k_{q,z}\leq n_{\max}-n_0 .
% \label{eq:allocation_objective}
% \end{equation}

\vspace{-2mm}

\begin{equation}
\max_{\{k_{q,z}\}}
\sum_{q\in\mathcal B}\sum_{z\in\mathcal A}
U_{q,z}(k_{q,z}), \quad
\sum_{q\in\mathcal B}\sum_{z\in\mathcal A}
k_{q,z}
\le
|\mathcal B|\, b_{\mathrm{extra}},
\quad
0\le k_{q,z}\le n_{\max}-n_0 .
\end{equation}

using greedy marginal gains
$\Delta_{q,z}(k)=U_{q,z}(k+1)-U_{q,z}(k)$.
Since $U_{q,z}$ is monotone and concave in integer $k$, this greedy allocation is optimal for the separable objective. We use $n_{\max}=8$ and $b_{\mathrm{extra}}=10$. With five strategies plus the unguided arm, this corresponds to 22 rollouts per prompt on average across the minibatch. Because $\varnothing$ competes under the same objective, guidance is selected only when it offers greater expected learning value than continued autonomous sampling.

% We use $n_{\max}=8$ and $B_{\mathrm{extra}}=10$. With five strategies plus the unguided arm, the main configuration therefore uses $22$ rollouts per prompt.

Both probe and additionally allocated samples form the same reward group and contribute to the arm-wise advantage whenever the resulting group is non-saturated.

\subsection{Learning from Strategy-Conditioned Rollouts}
\label{sec:guided_update}

All rollouts from the same $(q,z)$ arm form one reward group.
We compute within-arm advantages,
\vspace{-2mm}
\begin{equation}
A_i = \frac{r_i-\mu_{q,z}}
{\sigma_{q,z}+\epsilon},
\label{eq:arm_advantage}
\end{equation}
and discard groups with zero reward variance.
This avoids using a common baseline across strategy conditions whose success rates can differ
substantially.

For $z=\varnothing$, the update reduces to GRPO. For $z\neq\varnothing$, the rollout was sampled from
$\pi_{\theta_{\mathrm{old}}}(\cdot\mid c(q,z))$ while deployment uses
$c(q,\varnothing)$.
We transfer guided trajectory to unguided policy by an off-context importance correction.
% Let
% \begin{equation}
% w(y)
% =
% \frac{
% \pi_{\theta_{\mathrm{old}}}(y\mid c(q,\varnothing))
% }{
% \pi_{\theta_{\mathrm{old}}}(y\mid c(q,z))
% },
% \end{equation}
% be the sequence-level behavior correction, clipped in log space, and let
% \begin{equation}
% \rho^{\varnothing}_{\theta,u}
% =
% \frac{
% \pi_\theta(y_u\mid c(q,\varnothing),y_{<u})
% }{
% \pi_{\theta_{\mathrm{old}}}(y_u\mid c(q,\varnothing),y_{<u})
% },
% \end{equation}
Let the sequence-level behavior correction and token-level policy ratio be
\begin{equation}
w(y)
=
\frac{
\pi_{\theta_{\mathrm{old}}}(y\mid c(q,\varnothing))
}{
\pi_{\theta_{\mathrm{old}}}(y\mid c(q,z))
}, \quad
\rho^{\varnothing}_{\theta,u}
=
\frac{
\pi_\theta(y_u\mid c(q,\varnothing),y_{<u})
}{
\pi_{\theta_{\mathrm{old}}}(y_u\mid c(q,\varnothing),y_{<u})
}.
\label{eq:offcontext_ratios}
\end{equation}
Here $w(y)$ corrects the guided behavior distribution at the sequence level and is clipped in log space, while $\rho^{\varnothing}_{\theta,u}$ is the standard token-level PPO ratio under the unguided context. The transfer loss is

\vspace{-4mm}

\begin{equation}
\mathcal L_{\mathrm{tr}}
=
-\frac{1}{|y|}
\sum_u
\min\!\left(
\rho^{\varnothing}_{\theta,u}A_i\,w(y),\,
\operatorname{clip}
(\rho^{\varnothing}_{\theta,u},1\!\pm\!\epsilon)\,
A_i\,w(y)
\right).
\label{eq:transfer_loss}
\end{equation}

\vspace{-2mm}

This update uses strategy conditioning to obtain informative trajectories while optimizing the
unguided policy that is used at inference time.

\subsection{Maintaining Strategy Accessibility}
\label{sec:preservation}

The allocation mechanism assumes that the strategy instructions remain effective control handles.
Section~\ref{sec:strategy_analysis} shows that their steerability can decay during RL.
For a guided rollout, we therefore retain a small on-context PPO term
\begin{equation}
\mathcal L_{\mathrm{pres}}
=
-\frac{1}{|y|}
\sum_u
\min\!\left(
\rho^z_{\theta,u}A_i,\,
\operatorname{clip}
(\rho^z_{\theta,u},1\!\pm\!\epsilon)\,A_i
\right),
\qquad
\rho^z_{\theta,u}
=
\frac{
\pi_\theta(y_u\mid c(q,z),y_{<u})
}{
\pi_{\theta_{\mathrm{old}}}(y_u\mid c(q,z),y_{<u})
}.
\label{eq:preserve_loss}
\end{equation}
The final objective is
\begin{equation}
\mathcal L_{\mathrm{PSRA}}
=
\mathcal L_{\mathrm{tr}}
+
\lambda_t\,
\mathbb I[z\neq\varnothing]\,
\mathcal L_{\mathrm{pres}}.
\label{eq:psra_objective}
\end{equation}
The preservation term is applied only to guided arms.
Its coefficient is kept small and can be increased as the measured displacement
$D_t(z)$ declines.
Consequently, successful guided trajectories are transferred into autonomous behavior while the
strategy conditions that expose alternative reasoning routes remain usable during subsequent
allocation steps.

% \paragraph{Training loop.}
% Each PSRA step (i) probes all problem--strategy arms and updates their Beta posteriors, (ii) greedily allocates the remaining rollout budget using Eq.~\eqref{eq:allocation_objective}, and (iii) performs arm-wise policy updates using Eq.~\eqref{eq:psra_objective}.
% The strategy repertoire and DPMM coordinate system remain fixed throughout RL; only the posterior statistics, online opportunity estimates, and policy are updated.

\section{Experiment}

\subsection{Experimental Setup}

\paragraph{Models and Datasets.}
We conduct experiments on Qwen2.5-7B-Instruct, Qwen2.5-3B-Instruct, and Qwen2.5-1.5B-Instruct \citep{Yang2024Qwen25TR}. Unless otherwise specified, models are trained on DAPO-Math-17K-Processed \citep{Yu2025DAPOAO} with Math-Verify for correctness rewards. To examine robustness, we additionally train Qwen2.5-7B-Instruct on DeepScaleR-40K \citep{deepscaler2025}. 

\begingroup
\setlength{\tabcolsep}{4.2pt}
\renewcommand{\arraystretch}{1.12}

% Colors
\definecolor{sectiongray}{HTML}{E5E5E5}
\definecolor{sectionblue}{HTML}{DCE7FA}
\definecolor{avgyellow}{HTML}{FFF8CF}
\definecolor{avgblue}{HTML}{D9EEF5}
\definecolor{citegreen}{HTML}{00A51A}

\newcommand{\citeg}[1]{\textcolor{citegreen}{[#1]}}
\newcommand{\best}[1]{\textbf{#1}}
\newcommand{\second}[1]{\underline{#1}}

% Average columns
\newcommand{\avgid}[1]{%
  \cellcolor{avgyellow}#1%
}
\newcommand{\avgod}[1]{%
  \cellcolor{avgblue}#1%
}

% Section rows
\newcommand{\groupgray}[1]{%
  \multicolumn{11}{c}{\cellcolor{sectiongray}\strut #1}%
}
\newcommand{\groupblue}[1]{%
  \multicolumn{11}{c}{\cellcolor{sectionblue}\strut #1}%
}

\begin{table}[ht]
\centering
\caption{Main results across model scales, training dataset, and evaluation distributions.}
\label{tab:main_results}
\vspace{-2mm}

\resizebox{\textwidth}{!}{%
% 左右两端保留默认留白，使背景与整表横线边界对齐
\begin{tabular}{l*{10}{>{\centering\arraybackslash}c}}

\toprule

\multicolumn{1}{l}{\multirow{2}{*}{\textbf{Model}}}
& \multicolumn{6}{c}{\textbf{In-Distribution Performance}}
& \multicolumn{4}{c}{\textbf{Out-of-Distribution Performance}}
\\

% 取消第一条分组横线的右端裁切，
% 使其右端与黄色 Avg@8 列的右边界对齐
\cmidrule(l){2-7}
\cmidrule(l){8-11}

& \textbf{AIME 24/25}
& \textbf{AMC}
& \textbf{MATH-500}
& \textbf{Minerva}
& \textbf{OmniMath}
& \avgid{\textbf{Avg@8}}
& \textbf{StrategyQA}
& \textbf{GPQA}
& \textbf{MMLU-Pro}
& \avgod{\textbf{Avg@8}}
\\

\midrule

\groupblue{\textbf{Qwen2.5-7B/3B/1.5B-Instruct on DAPO-Math-17K}}
\\

\midrule

\rowcolor{gray!18}
Qwen2.5-7B-Instruct
& 10.2/6.9 & 51.8 & 74.4 & 35.5 & 24.6 & \avgid{33.9}
& 63.0 & 27.3 & 23.8 & \avgod{38.0}
\\

+Vanilla GRPO
& 10.8/6.9 & 53.3 & 75.2 & 36.7 & 25.1 & \avgid{34.6}
& 63.0 & 27.0 & 22.1 & \avgod{37.4}
\\

LUFFY
& 12.2/6.6 & 54.2 & 75.1 & 36.3 & 25.3 & \avgid{35.0}
& 63.4 & 29.6 & 25.4 & \avgod{39.5}
\\

BREAD
& 12.6/7.7 & 54.2 & 74.4 & 35.6 & 24.6 & \avgid{34.9}
& 63.0 & 30.0 & 24.2 & \avgod{39.1}
\\

HINT
& 12.6/8.2 & 53.3 & 75.0 & \best{39.2} & 25.6 & \avgid{35.7}
& 55.6 & 34.2 & 26.0 & \avgod{38.6}
\\

+Fixed Strategy
& 11.6/7.7 & 53.3 & 75.1 & 36.9 & 25.1 & \avgid{34.9}
& 61.8 & 33.3 & \best{26.1} & \avgod{40.3}
\\

\textbf{+PSRA}
& \best{12.8/9.8}
& \best{57.6}
& \best{76.5}
& \second{37.9}
& \best{26.6}
& \avgid{\best{36.9}}
& \best{64.3}
& \best{34.7}
& 23.2
& \avgod{\best{40.7}}
\\

\midrule

\rowcolor{gray!18}
Qwen2.5-3B-Instruct
& 6.0/2.5 & 38.3 & 65.1 & 27.3 & 19.1 & \avgid{26.4}
& 49.1 & 31.2 & \best{37.4} & \avgod{39.2}
\\

+Vanilla GRPO
& \best{6.7}/3.1 & 41.0 & 65.8 & 27.6 & 19.2 & \avgid{27.2}
& 50.1 & 30.7 & 35.9 & \avgod{38.9}
\\

\textbf{+PSRA}
& \second{6.5}/\best{3.3}
& \best{42.6}
& \best{67.9}
& \best{28.8}
& \best{19.6}
& \avgid{\best{28.1}}
& \best{59.9}
& \best{33.9}
& \second{37.1}
& \avgod{\best{43.6}}
\\

\midrule

\rowcolor{gray!18}
Qwen2.5-1.5B-Instruct
& 3.3/1.5 & 27.9 & 52.4 & 16.9 & 14.1 & \avgid{19.3}
& 0.1 & 14.9 & 8.3 & \avgod{7.8}
\\

+Vanilla GRPO
& 2.7/1.3 & 30.0 & 54.4 & 18.3 & 14.5 & \avgid{20.2}
& 0.5 & 15.1 & \best{8.9} & \avgod{8.2}
\\

\textbf{+PSRA}
& \best{3.6/3.3}
& \best{32.3}
& \best{56.2}
& \best{19.5}
& \best{16.2}
& \avgid{\best{21.9}}
& \best{2.1}
& \best{15.9}
& 7.9
& \avgod{\best{8.6}}
\\

\midrule

\groupblue{\textbf{Qwen2.5-7B-Instruct on DeepScaleR-40k}}
\\

\midrule

\rowcolor{gray!18}
+Vanilla GRPO
& 11.2/5.8 & 52.5 & 74.7 & 36.2 & 25.2 & \avgid{34.3}
& 65.4 & \best{28.2} & 25.8 & \avgod{39.8}
\\

+Fixed Strategy
& 9.6/5.6 & 52.9 & 75.9 & 36.3 & 26.0 & \avgid{34.4}
& 66.6 & 27.3 & 26.7 & \avgod{40.2}
\\

\textbf{+PSRA}
& \best{12.9/7.2}
& \best{53.6}
& \best{76.7}
& \best{36.9}
& \best{26.5}
& \avgid{\best{35.6}}
& \best{67.5}
& \second{27.1}
& \best{29.8}
& \avgod{\best{41.5}}
\\

\bottomrule

\end{tabular}%
}

\vspace{-4mm}
\end{table}
\endgroup

\paragraph{Evaluation Benchmarks and Metrics.}
We evaluate in-distribution (ID) mathematical reasoning capability on AIME 2024/2025, AMC 2023, MATH-500 \citep{lightman2023let}, Minerva Math \citep{lewkowycz2022minerva}, and Omni-MATH \citep{Gao2024OmniMATHAU}.  To test whether the learned reasoning improvements transfer beyond the mathematical training domain, we further evaluate out-of-distribution (OOD) generalization on StrategyQA \citep{geva2021strategyqa}, GPQA-Diamond \citep{rein2023gpqa}, and MMLU-Pro \citep{wang2024mmlupro}, which emphasize implicit logical reasoning and broad knowledge-intensive reasoning. In the main results, we report \textsc{Pass@1} and \textsc{Avg@8}. To measure reasoning coverage under larger inference budgets, we additionally report \textsc{Pass@$k$} for $k$ up to 128.

\paragraph{Baselines and Ours.} We compare against \textit{Vanilla GRPO} \citep{DeepSeekAI2025DeepSeekR1IR}, \textit{LUFFY} \citep{yan2025luffy}, \textit{BREAD} \citep{Zhang2025BREADBR}, and \textit{HINT} \citep{wang2026hint}, representing standard on-policy RLVR and representative guided or off-policy RLVR methods. We additionally consider \textit{Fixed Strategy}, which uses the random $5$ strategy anchors to isolate the effect of the inner discovered strategy. All methods are matched to the same total generation budget, equivalent to 22 policy rollouts per training prompt on average. Refer to Appendix~\ref{app:training_details} for our training details.

\subsection{Main Results}
\begin{figure}
    \centering
    \includegraphics[width=1\linewidth]{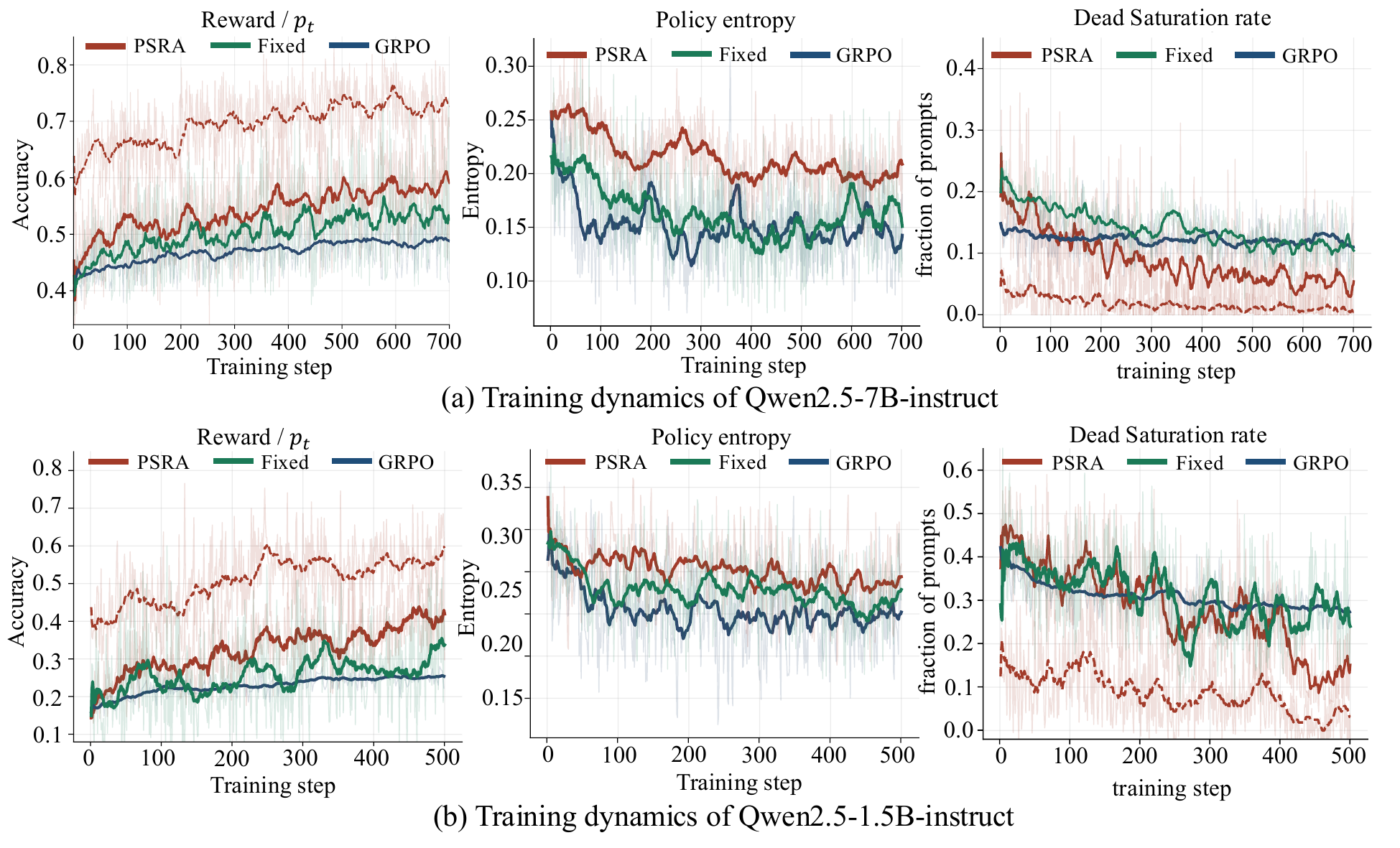}
    \vspace{-6mm}
    \caption{\textbf{Training dynamics on Qwen2.5-7B-Instruct and Qwen2.5-1.5B-Instruct.} Under the same rollout budget, PSRA achieves faster and more stable reward improvement, maintains higher policy entropy, and reduces dead-saturated groups. The dashed curve denotes the best strategy arm, illustrating the remaining strategy-conditioned performance frontier.}
    \label{fig:training_dynamics}
    \vspace{-6mm}
\end{figure}

Table~\ref{tab:main_results} shows that PSRA consistently improves in-distribution reasoning across training corpora and model scales. On Qwen2.5-7B-Instruct, PSRA outperforms vanilla GRPO by 2.3 points on DAPO-Math-17K and 1.3 points on DeepScaleR-40K, while exceeding the strongest competing baseline by 1.2 points in both settings.
The gains persist at smaller scales, with improvements of 0.9 and 1.7 points over vanilla GRPO on the 3B and 1.5B models. The larger gain on 1.5B is particularly notable, since smaller models exhibit substantially higher dead-saturation rates under vanilla GRPO (Figure~\ref{fig:training_dynamics}); by switching toward alternative strategies already accessible to the model, PSRA recovers informative training signals on otherwise saturated problems.

\begin{figure}
    \centering
    \includegraphics[width=1\linewidth]{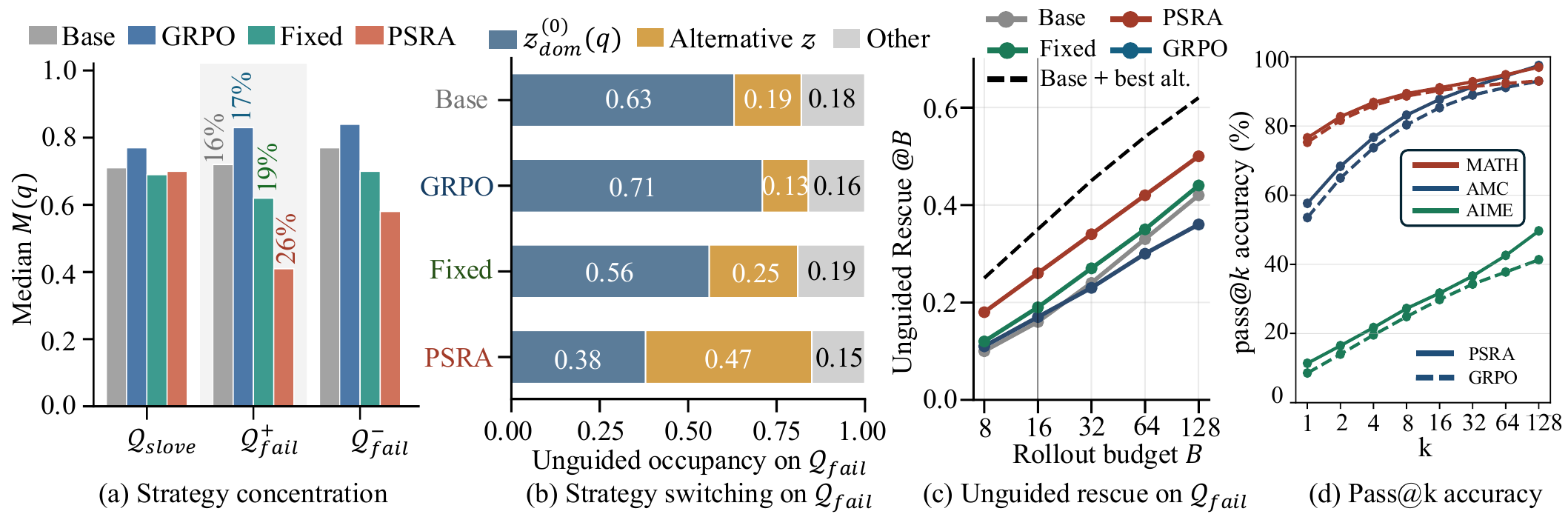}
    \vspace{-2mm}
    \caption{\textbf{Strategy-level effects of PSRA.}
(a) PSRA successfully reduces strategy concentration.
(b) Its unguided policy shifts probability mass from the base-dominant strategy toward alternative strategies.
(c) This produces higher rescue rates on initially failed problems across rollout budgets.
(d) PSRA maintains a growing advantage over vanilla GRPO at larger inference budgets.}
    \label{fig:strategy}
    \vspace{-5mm}
\end{figure}

PSRA also yields more consistent OOD transfer. While vanilla GRPO shows mixed behavior outside the mathematical training distribution, PSRA improves the OOD average across all settings, outperforming GRPO by 3.3, 4.7, and 0.4 points on the 7B, 3B, and 1.5B models trained on DAPO-Math-17K, and by 1.7 points on the 7B model trained on DeepScaleR-40K. This suggests that problem-conditioned strategy switching promotes a broader and more transferable reasoning repertoire. Moreover, Figure~\ref{fig:strategy}d shows that PSRA's advantage generally widens as the inference budget $k$ increases, especially on AIME, indicating that the method improves not only single-sample accuracy but also preserves broader strategy coverage under larger test-time budgets.

\vspace{-2mm}

\paragraph{Reasoning strategy analysis.}
We examine whether PSRA changes the strategy selected by the unguided policy. Let $\mathcal Q_{\rm fail}$ denote held-out problems for which all $B{=}16$ base-model rollouts fail, and split them into $\mathcal Q_{\rm fail}^{+}$ and $\mathcal Q_{\rm fail}^{-}$ according to whether each trained policy subsequently solves them. PSRA recovers 26\% of $\mathcal Q_{\rm fail}$, compared with 17\% for GRPO and 19\% for \textit{Fixed Strategy} (Fig.~\ref{fig:strategy}a,c). More importantly, its recovered problems are less concentrated on the base-dominant strategy, and 73\% are first solved within an alternative strategy footprint, while strategy concentration remains similar across methods on $\mathcal Q_{\rm solve}$. This indicates problem-conditioned switching rather than a uniform increase in diversity. Consistently, the unguided PSRA policy shifts substantially more mass toward admitted alternatives and approaches the base model's guided best-alternative rescue rate across budgets (Fig.~\ref{fig:strategy}b,c). Finally, Fig.~\ref{fig:analysis}g shows that under PSRA, reductions in conditional displacement are largely explained by internalization $\iota_t$, whereas GRPO often loses strategy accessibility before absorption. Together, these results suggest that PSRA successfully internalized strategy-conditioned behaviors into autonomous reasoning.

% \paragraph{Training Dynamics}
% Figure~\ref{fig:training_dynamics} shows that, under the same rollout budget, PSRA learns faster and more stably while maintaining substantially higher policy entropy, indicating more sustained exploration.
% Its dead-saturation rate quickly falls below GRPO on the 7B model; the 1.5B model adapts more slowly and exhibits a brief initial increase in saturation, but eventually achieves the same advantage. This transient behavior is consistent with smaller models requiring more updates to adjust their problem-conditioned strategy-selection prior after switching among already available reasoning strategies. Meanwhile, the best-strategy arm (dashed) remains consistently strong, while the unguided policy steadily improves toward it. These complementary trends reflect the two roles of our objective: the preservation term keeps useful strategy-conditioned routes callable, whereas the transfer term progressively internalizes their successful behaviors into the unguided policy. Moreover, selecting the best of multiple finite-sample strategy arms introduces an optimistic noise floor of approximately $0.20$ in our setting; after accounting for this effect, the remaining gap is small, suggesting that much of the strategy-specific gain has been absorbed into autonomous reasoning.

\paragraph{Training Dynamics.}
Figure~\ref{fig:training_dynamics} shows that PSRA learns faster and more stably under the same rollout budget, while maintaining higher policy entropy and lower dead-saturation rates than GRPO. The reduction in saturation appears earlier on 7B than on 1.5B, suggesting that smaller models adapt more slowly to problem-conditioned strategy switching, but ultimately benefit from the same mechanism. Meanwhile, the best-strategy arm remains strong as the unguided policy steadily approaches it, consistent with preservation keeping useful routes callable while transfer progressively internalizes them. After accounting for the $\approx 0.20$ optimistic noise floor, the remaining gap is small, indicating that much of the strategy-specific gain has been absorbed into autonomous reasoning.

\subsection{Ablation Studies}

\begin{wrapfigure}{r}{0.42\textwidth}
    \centering
    \includegraphics[width=0.44\textwidth]{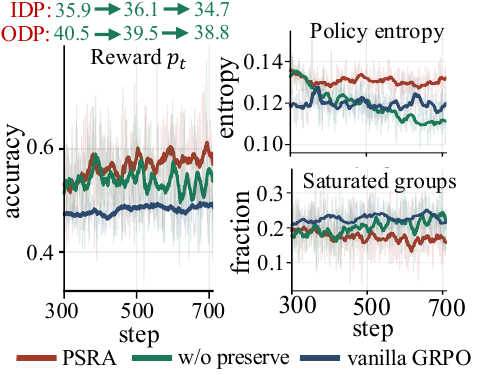}
    \vspace{-4mm}
    \caption{Strategy preservation ablation.}
    \label{fig:ablate_perserve}
    \vspace{-5mm}
\end{wrapfigure}

\paragraph{Fixed Guidance.}
We first replace the discovered strategy set with a fixed set of randomly selected guidance prompts (see Appendix~\ref{app:fixed_strategy}). While such guidance can improve over vanilla GRPO in some settings, it consistently underperforms PSRA (Table~\ref{tab:main_results}) and recovers only 19\% of initially unsolved problems, compared with 26\% for PSRA (Figure~\ref{fig:strategy}). Figure~\ref{fig:training_dynamics} further shows higher saturation and weaker training progress, indicating that guidance alone is insufficient: effective exploration depends on steering toward strategy that are both accessible to the model and useful for the problem.

% \begin{wraptable}{r}{0.40\textwidth}
%     \vspace{-10pt}   
%     \centering
%     \setlength{\tabcolsep}{5pt}
%     \small
%     \caption{Ablation of strategy-set size and Bayesian rollout allocation.}
%     \label{tab:ablation}
%     \begin{tabular}{cccc}
%         \toprule[1pt]
%         & \textbf{Variant} & \textbf{IDP Avg.} & \textbf{ODP Avg.} \\
%         \midrule[0.8pt]
%         \multirow{4}{*}{\rotatebox{90}{\textit{strategy}}}   
%         & \#4  & 36.3$_{\pm \text{0.4}}$ & 40.3$_{\pm \text{0.3}}$ \\
%         & \#3  & 36.1$_{\pm \text{0.3}}$ & 40.3$_{\pm \text{0.4}}$ \\
%         & \#2  & 34.6$_{\pm \text{0.6}}$ & 38.0$_{\pm \text{0.8}}$ \\
%         & \#1  & 35.3$_{\pm \text{0.5}}$ & 38.1$_{\pm \text{0.6}}$ \\
%         \midrule[0.5pt]
%         \multirow{5}{*}{\rotatebox{90}{\textit{bayes}}}
%         & Rand   & 35.6$_{\pm \text{0.5}}$ & 38.2$_{\pm \text{0.6}}$ \\
%         & $B0$   & 35.9$_{\pm \text{0.7}}$ & 38.0$_{\pm \text{0.5}}$ \\
%         & $B5$   & 36.3$_{\pm \text{0.7}}$ & 38.8$_{\pm \text{0.5}}$ \\
%         & $B15$  & 36.9$_{\pm \text{0.5}}$ & 40.3$_{\pm \text{0.9}}$ \\
%         & $B20$  & 37.2$_{\pm \text{0.6}}$ & 40.7$_{\pm \text{0.6}}$ \\
%         \bottomrule[1pt]
%     \end{tabular}
%     \vspace{-10pt}  
% \end{wraptable}

\vspace{-2mm}

\paragraph{Number of Strategies.}
Table~\ref{tab:ablation} shows that PSRA is relatively insensitive to moderate reductions in the strategy set: using four or three strategies yields performance close to the full configuration.
Performance drops sharply with only two strategies, even falling below the single-strategy variant, suggesting that a small set of competing but insufficiently complementary directions can interfere with effective routing.
Overall, the results support using a compact but sufficiently diverse strategy basis rather than maximizing the number of guidance arms.

\vspace{-2mm}

\paragraph{Bayesian Allocation.}
Randomly assigning the additional rollouts performs worst among the allocation variants and even underperforms $B{=}0$ on ID performance, despite using more rollouts, showing that poorly targeted compute can introduce harmful training signals. In contrast, Bayesian allocation improves steadily as additional budget is introduced; increasing the budget beyond the default $B{=}10$ yields only marginal further gains while incurring proportionally higher rollout cost.

\vspace{-2mm}

\paragraph{Strategy Preservation.}
Figure~\ref{fig:ablate_perserve} further isolates the preservation objective.
Removing it leads to faster entropy decay, more saturated groups, and weaker final performance, while full PSRA maintains stronger exploration and more stable optimization.
This supports preservation as a necessary component for keeping strategy routes accessible to be transferred into autonomous reasoning.

\vspace{-2mm}

\begin{table*}[t]
    \centering
    \footnotesize
    \setlength{\tabcolsep}{5pt}
    \caption{Ablation of strategy-set size and Bayesian rollout allocation. $B$ denotes reasoning budget.}
    \vspace{-2mm}
    \label{tab:ablation}
    \begin{tabular}{lcccc@{\hspace{5pt}}ccccc}
        \toprule
        & \multicolumn{4}{c}{\textit{Number of strategies}}
        & \multicolumn{5}{c}{\textit{Bayes allocation budget}} \\
        \cmidrule(lr){2-5}
        \cmidrule(lr){6-10}
        \textbf{Variant}
        & \#4 & \#3 & \#2 & \#1
        & Rand & $B0$ & $B5$ & $B15$ & $B20$ \\

        \textbf{IDP Avg.}
        & $36.3_{\pm.4}$
        & $36.1_{\pm.3}$
        & $34.6_{\pm.6}$
        & $35.3_{\pm.5}$
        & $35.6_{\pm.5}$
        & $35.9_{\pm.7}$
        & $36.3_{\pm.7}$
        & $36.9_{\pm.5}$
        & $\mathbf{37.2}_{\pm.6}$ \\

        \textbf{ODP Avg.}
        & $40.3_{\pm.3}$
        & $40.3_{\pm.4}$
        & $38.0_{\pm.8}$
        & $38.1_{\pm.6}$
        & $38.2_{\pm.6}$
        & $38.0_{\pm.5}$
        & $38.8_{\pm.5}$
        & $40.3_{\pm.9}$
        & $\mathbf{40.7}_{\pm.6}$ \\
        \bottomrule
    \end{tabular}
    \vspace{-2mm}
\end{table*}

\section{Discussion and Limitations}
Our experiments span multiple model scales and training corpora, but remain limited to the Qwen2.5 family up to 7B; broader model families and larger models remain important directions for future validation.
Nevertheless, the shared strategy families observed across scales suggest that strategy conditioning provides a stable interface for RLVR exploration, while adapting PSRA to a new model requires only lightweight strategy discovery and calibration on a small sample relative to the cost of RL training. For complete examples of how PSRA improves reasoning through strategy switching, please refer to Appendix~\ref{app:strategy_examples}. PSRA also introduces little online overhead and reaches comparable reward levels in markedly fewer training steps while maintaining stronger exploration (Figure~\ref{fig:training_dynamics} ).

\vspace{-2mm}

\section{Conclusion}
We show that RLVR exploration failures can arise from finite-budget strategy selection rather than missing capability. Motivated by this view, we introduce PSRA, which adaptively allocates rollouts over problem--strategy arms while preserving useful strategy conditions and transferring successful guided behaviors into the unguided policy. Across model scales and training settings, PSRA improves reasoning performance, reduces saturation, and sustains broader exploration, highlighting pretrained strategy repertoires as an effective interface for RLVR exploration.

% \section{AI use statement}
% We used LLMs to assist with writing and polishing the manuscript and to support retrieval and discovery, such as identifying related work. All AI-assisted content and suggested references were reviewed and verified. We take full responsibility for the final content of this paper, including the accuracy and integrity of its text, claims, and references.

% \section{Reproducibility statement}
% To support the reproducibility of our work, we provide a detailed description of the proposed method in the main paper and have submitted the code as supplementary material. We have also documented the experimental setup, including the datasets, evaluation protocols, and implementation details, in the paper and the supplementary material.

\bibliography{iclr2027_conference}
\bibliographystyle{iclr2027_conference}

\appendix
\section{Appendix}

\subsection{Detailed Strategy Discovery and Problem-Conditioned Exploration}
\label{app:strategy_analysis}

This appendix provides additional details for the strategy discovery, behavioral certification, problem-conditioned rescue analysis, coverage predictor, and longitudinal accessibility analysis introduced in Section~\ref{sec:strategy_analysis}. Unless otherwise stated, all representation-space analyses use the frozen base-model coordinate system constructed before RL training.

\subsubsection{Strategy Repertoire Discovery}
\label{app:strategy_discovery}

\paragraph{Rollout representations.}
We first construct a repertoire discovery set $D_{\mathrm{rep}}$ consisting of problems and multiple autonomous rollouts generated by the frozen base policy.
For every rollout $y$, we re-encode the complete problem--response sequence under the original problem context and extract hidden states from early and intermediate portions of the generated response.
Token-level states within each region are pooled and concatenated to obtain a rollout-level representation
\begin{equation}
    e_{\mathrm{s}}(y)\in\mathbb{R}^{d}.
\end{equation}
All subsequent clustering and longitudinal measurements use this representation extractor frozen at the base checkpoint, preventing changes in the representation space itself from being confounded with changes in policy behavior.

\paragraph{DPMM clustering and semantic merging.}
We fit a Dirichlet process mixture model (DPMM) to
$\{e_{\mathrm{s}}(y):y\in D_{\mathrm{rep}}\}$.
Unlike a fixed-$K$ clustering method, the DPMM allows the number of fine-grained components to be determined from the rollout distribution.
Let
\begin{equation}
    c(y)\in\{1,\ldots,K_{\mathrm{DPMM}}\}
\end{equation}
denote the frozen component assignment of rollout $y$.
We inspect representative correct trajectories from stable components and merge components exhibiting the same high-level solution procedure into semantic strategy candidates.
This procedure yields seven candidates, five of which survive the behavioral certification described below and form the final strategy set $\mathcal Z$.

Importantly, the DPMM components themselves remain the fine-grained coordinate system throughout the paper.
Semantic strategies are defined as unions of these components rather than as replacement cluster labels; this allows the same representation space to support both strategy-level analysis and finer-grained coverage estimation.

\subsubsection{Behavioral Certification and Strategy Footprints}
\label{app:strategy_certification}

Representation-space coherence alone does not establish that a cluster corresponds to a controllable reasoning strategy.
We therefore behaviorally certify each candidate on a disjoint calibration set $D_{\mathrm{cal}}$ using three conditions:

\begin{enumerate}
    \item \textbf{Unguided}: the original problem prompt, denoted by $\varnothing$;
    \item \textbf{Exemplar-conditioned}: the prompt is augmented with representative trajectories associated with candidate strategy $z$;
    \item \textbf{Instruction-conditioned}: the exemplars are replaced by a short natural-language description of the same strategy.
\end{enumerate}

We additionally use matched placebo instructions that preserve prompt length and instructional form without specifying the target reasoning procedure.
A candidate is retained only when both exemplar and instruction conditioning consistently increase occupancy of its associated representation region relative to the unguided and placebo controls.
This test distinguishes a genuinely steerable reasoning direction from a cluster that is merely descriptively coherent.

Let $\mathcal K_z$ be the set of DPMM components semantically associated with strategy $z$.
For each component, we measure the intervention lift
\begin{equation}
    \ell_z(k)
    =
    \frac{
        P(c(y)=k\mid z)
    }{
        P(c(y)=k\mid\varnothing)
    },
    \qquad k\in\mathcal K_z .
    \label{eq:app_intervention_lift}
\end{equation}
We rank components in $\mathcal K_z$ by $\ell_z(k)$ and define the strategy footprint $C_z$ as the smallest prefix of this ranking whose cumulative probability accounts for at least $50\%$ of the strategy-conditioned mass within $\mathcal K_z$.
The footprint therefore contains the components most causally associated with invoking $z$, rather than all components that happen to share its semantic label.

Natural-language instructions retain most of the representation displacement produced by exemplar conditioning, whereas matched placebo instructions produce negligible displacement (Fig.~\ref{fig:analysis}c).
We consequently use natural-language instructions as the lightweight strategy handles during RL.

\subsubsection{Problem-Conditioned Concentration and Rescue}
\label{app:strategy_rescue}

For each problem $q$, we quantify how strongly autonomous sampling occupies the footprint of strategy $z$:
\begin{equation}
    p_{\varnothing}(z\mid q)
    =
    P\!\left(c(y)\in C_z\mid q,\varnothing\right).
    \label{eq:app_strategy_occupancy}
\end{equation}
Because a rollout may fall outside every admitted strategy footprint, we additionally normalize occupancy over the strategy-associated mass,
\begin{equation}
    \widetilde p_{\varnothing}(z\mid q)
    =
    \frac{
        p_{\varnothing}(z\mid q)
    }{
        \sum_{z'\in\mathcal Z}p_{\varnothing}(z'\mid q)
    },
\end{equation}
and define the dominant autonomous strategy as
\begin{equation}
    z_{\mathrm{dom}}(q)
    =
    \arg\max_{z\in\mathcal Z}
    p_{\varnothing}(z\mid q).
    \label{eq:app_dominant_strategy}
\end{equation}

\paragraph{Rescue protocol.}
To determine whether an all-failure rollout group necessarily indicates missing capability, we isolate problems
\begin{equation}
    \mathcal Q_{\mathrm{fail}}
    =
    \left\{
    q:
    r(y_i)=0,\ \forall i\le B_{\mathrm{init}}
    \right\},
\end{equation}
for which the initial autonomous rollout group contains no correct response.
Given an additional budget $B$, we define
\begin{equation}
    \mathrm{Rescue@}B(a)
    =
    \Pr_{q\in\mathcal Q_{\mathrm{fail}}}
    \left[
        \exists\, i\le B:
        r\!\left(y_i^{(a)}\right)=1
    \right],
    \label{eq:app_rescue}
\end{equation}
where $a$ denotes the sampling condition.

We compare continued unguided sampling, sampling under the base-dominant strategy $z_{\mathrm{dom}}(q)$, higher-temperature unguided sampling, and an oracle best-alternative strategy.
The latter is defined only for analysis as
\begin{equation}
    z_{\mathrm{alt}}^{*}(q)
    =
    \arg\max_{z\in\mathcal Z\setminus
    \{z_{\mathrm{dom}}(q)\}}
    P(r=1\mid q,z).
    \label{eq:app_oracle_alt}
\end{equation}
Importantly, $z_{\mathrm{alt}}^{*}$ is computed from evaluation rollouts and is \emph{never} available to PSRA during training.
It provides an oracle estimate of the headroom available from strategy switching.

Figure~\ref{fig:analysis}d reports Rescue@$B$ over increasing budgets.
The substantially higher rescue rate achieved by the oracle alternative shows that many all-failure groups arise from finite-budget strategy concentration rather than from a complete absence of successful behavior in the base policy.

\subsubsection{Training the Pre-Generation Coverage Predictor}
\label{app:coverage_predictor}

Section~\ref{sec:strategy_analysis} introduces a predictor
\begin{equation}
    G:
    h_{\mathrm{plan}}(q)
    \rightarrow
    \mathrm{PCA}_{128}
    \rightarrow
    \mathrm{MLP}_{128\rightarrow64\rightarrow K_{\mathrm{DPMM}}}
    \rightarrow
    \widehat q(c\mid q),
    \label{eq:app_predictor_arch}
\end{equation}
which estimates the distribution of reasoning components that may be reached from problem $q$ before any response tokens are generated.

\paragraph{Training data.}
For every training problem $q$, we extract the hidden state
$h_{\mathrm{plan}}(q)$ at the final prompt position from the frozen base model.
We then independently sample multiple continuations
\[
    Y_q=\{y_{q,1},\ldots,y_{q,n_q}\}
\]
and assign every rollout to its frozen DPMM component $c(y)$.
Thus, all trajectories for a given problem share the same predictor input but provide independent samples from the future reasoning distribution induced by that problem.

Problems used to fit $G$ are disjoint from the analysis set
$D_{\mathrm{eval}}$ used to report Figure~\ref{fig:analysis}e.
The PCA projection is fitted using training problems only and then frozen together with the MLP for all subsequent evaluations and RL experiments.

\paragraph{Training objective.}
We train $G$ by maximum likelihood over rollout-level DPMM assignments:
\begin{equation}
    \mathcal L_G
    =
    -
    \sum_{q\in D_G^{\mathrm{train}}}
    \frac{1}{|Y_q|}
    \sum_{y\in Y_q}
    \log
    \widehat q\!\left(c(y)\mid q\right).
    \label{eq:app_predictor_loss}
\end{equation}
Equivalently, this minimizes the cross-entropy between
$\widehat q(\cdot\mid q)$ and the empirical component-frequency distribution of future rollouts for $q$.
No rollout outcome from the finite observed set used later by PSRA is provided as input to $G$; the prediction is based only on the pre-generation problem representation.

We select model checkpoints using held-out prediction likelihood.
The early-layer baseline used in Figure~\ref{fig:analysis}e follows the same PCA--MLP architecture and training objective, differing only in the representation supplied to the predictor.
This controls for predictor capacity and isolates the information contained in the planning representation.

\paragraph{Novelty-weighted evaluation.}
For each held-out problem $q\in D_{\mathrm{eval}}$, we partition independently generated trajectories into a small observed set $O_q$ and a larger held-out set $H_q$.
For each $y\in H_q$, novelty relative to the finite observed group is
\begin{equation}
    d_{\min}(y,O_q)
    =
    \min_{y'\in O_q}
    d\!\left(
        e_{\mathrm{s}}(y),
        e_{\mathrm{s}}(y')
    \right),
\end{equation}
where $d(\cdot,\cdot)$ is the representation-space distance used throughout the analysis.
We convert these distances to rank-normalized weights
\begin{equation}
    w(y)
    =
    \operatorname{ranknorm}
    \left(d_{\min}(y,O_q)\right).
\end{equation}
The coverage score is then
\begin{equation}
    \mathrm{CovScore}(G)
    =
    \frac{
        \sum_{q\in D_{\mathrm{eval}}}
        \sum_{y\in H_q}
        w(y)
        \log
        \widehat q(c(y)\mid q)
    }{
        \sum_{q\in D_{\mathrm{eval}}}
        \sum_{y\in H_q}
        w(y)
    }.
    \label{eq:app_covscore}
\end{equation}
Higher-novelty trajectories receive larger weight, so this metric emphasizes the quality of $G$ specifically on reasoning directions that are poorly represented by a finite autonomous sample.

This evaluation differs from ordinary component-prediction accuracy in an important way: a predictor cannot obtain a high score merely by assigning most probability to the dominant reasoning mode.
It must also assign probability to plausible but under-sampled regions of the rollout distribution.

\subsubsection{From Predicted Coverage to Strategy Opportunities}
\label{app:coverage_mapping}

Given a finite observed rollout set $O_q$, let
\begin{equation}
    \mathrm{hit}(O_q)
    =
    \{c(y):y\in O_q\}
\end{equation}
be the set of DPMM components already observed.
We define the coverage tail
\begin{equation}
    U_m(q;O_q)
    =
    \operatorname{Top}_m
    \left\{
        \widehat q(k\mid q):
        k\notin\mathrm{hit}(O_q)
    \right\}
    \label{eq:app_coverage_tail}
\end{equation}
as the $m$ highest-probability predicted components that remain unvisited.

The strategy footprints then convert this representation-level prediction into executable exploration directions:
\begin{equation}
    \Omega(q,z;O_q)
    =
    \sum_{k\in U_m(q;O_q)\cap C_z}
    \widehat q(k\mid q).
    \label{eq:app_omega}
\end{equation}
Thus, $\Omega(q,z;O_q)$ is large only when (i) the model predicts substantial probability mass in currently unobserved regions and (ii) those regions are reachable through strategy $z$.

\paragraph{Strategy coverage analysis.}
To evaluate how much of the predicted coverage tail is addressable by the discovered repertoire, we greedily add certified strategies according to the incremental gap mass covered by their footprints.
For a strategy subset $\mathcal S\subseteq\mathcal Z$, define
\begin{equation}
    \mathrm{GapCov}(\mathcal S)
    =
    \frac{
        \sum_q
        \sum_{k\in U_m(q;O_q)}
        \widehat q(k\mid q)
        \mathbb I
        \left[
            k\in\bigcup_{z\in\mathcal S}C_z
        \right]
    }{
        \sum_q
        \sum_{k\in U_m(q;O_q)}
        \widehat q(k\mid q)
    }.
    \label{eq:app_gapcov}
\end{equation}
At each step we add the strategy with the largest marginal increase in
$\mathrm{GapCov}$.
We compare this curve against matched random subsets of DPMM components.
As shown in Figure~\ref{fig:analysis}f, coverage rises rapidly and saturates with a small number of certified strategies, showing that the discovered repertoire captures most of the structured coverage opportunities rather than merely partitioning representation space arbitrarily.

\subsubsection{Longitudinal Strategy Accessibility}
\label{app:accessibility}

To distinguish successful strategy internalization from loss of steerability, all longitudinal measurements are performed in the frozen base-model coordinate system.
In particular, the initial DPMM and strategy footprints
$C_z^{(0)}$ are never re-clustered after RL begins.

At checkpoint $t$, we generate matched unguided and strategy-conditioned rollouts and assign them to the frozen DPMM components.
We define the conditional displacement of strategy $z$ as
\begin{equation}
    D_t(z)
    =
    \mathrm{JS}
    \left(
        P_t(c\mid z)
        \,\Vert\,
        P_t(c\mid\varnothing)
    \right),
    \label{eq:app_displacement}
\end{equation}
and normalize its reduction relative to the base checkpoint:
\begin{equation}
    \delta_t(z)
    =
    1-\frac{D_t(z)}{D_0(z)}.
    \label{eq:app_displacement_reduction}
\end{equation}

A reduction in displacement is ambiguous by itself.
It may occur because autonomous generation has moved into the strategy region, which is desirable internalization, or because the strategy instruction has simply stopped changing the model's behavior.
We therefore measure autonomous occupancy of the initial footprint,
\begin{equation}
    \iota_t(z)
    =
    \frac{
        P_t(C_z^{(0)}\mid\varnothing)
        -
        P_0(C_z^{(0)}\mid\varnothing)
    }{
        P_0(C_z^{(0)}\mid z)
        -
        P_0(C_z^{(0)}\mid\varnothing)
    },
    \label{eq:app_internalization}
\end{equation}
and define residual accessibility loss as
\begin{equation}
    L_t(z)
    =
    \delta_t(z)-\iota_t(z).
    \label{eq:app_accessibility_loss}
\end{equation}

When $\delta_t(z)\approx\iota_t(z)$, the shrinking conditional gap is largely explained by the unguided policy moving into the region previously exposed by $z$.
We interpret this as \emph{internalization}.
In contrast, a large positive $L_t(z)$ indicates that strategy-conditioned displacement has disappeared without a corresponding increase in autonomous occupancy, which we interpret as \emph{loss of conditional accessibility}.
Figure~\ref{fig:analysis}g reports these two dynamics across RL checkpoints.

\paragraph{Connection to PSRA.}
These measurements play two different roles.
The frozen footprints and the coverage predictor provide the representation-space interface used to estimate problem--strategy opportunity during allocation.
The accessibility analysis additionally motivates the preservation objective in Section~\ref{sec:method}: a strategy that remains useful for exploration must stay callable long enough for its successful behaviors to be transferred to the unguided policy.
No re-clustering of the reasoning space is performed during RL; only policy-dependent occupancy, success statistics, and strategy accessibility evolve.

\subsubsection{Online Estimation of the PSRA Prior}
\label{app:online_prior}

We estimate the hierarchical prior in Eq.~(8) using only the all-arm probe.
Adaptively allocated rollouts are never used to update the prior statistics, preventing
the allocation policy from reinforcing its own previous choices.

\paragraph{Leave-one-arm-out problem effect.}
For target arm $z$, let
\[
S_{q,-z}
=
\sum_{a\in\mathcal A\setminus\{z\}} s_{q,a}.
\]
We estimate the current difficulty of problem $q$ using the probe outcomes of all
other arms with Jeffreys smoothing:
\begin{equation}
\hat c_t^{-z}(q)
=
\operatorname{logit}\!\left(
\frac{S_{q,-z}+\frac12}
{(|\mathcal A|-1)n_0+1}
\right).
\tag{41}
\end{equation}
Excluding $s_{q,z}$ prevents the outcome used in the posterior update of arm
$(q,z)$ from also entering its prior mean.

\paragraph{Global arm effect.}
Let
\[
\ell_t(q,z)
=
\operatorname{logit}\!\left(
\frac{s_{q,z}+\frac12}{n_0+1}
\right).
\]
For each arm, we compute its mean residual success relative to problem difficulty,
\begin{equation}
r_t(z)
=
\frac{1}{|\mathcal B|}
\sum_{q\in\mathcal B}
\left[
\ell_t(q,z)-\hat c_t^{-z}(q)
\right],
\tag{42}
\end{equation}
and update its global effect by an exponential moving average,
\begin{align}
\tilde\beta_{t+1}(z)
&=
(1-\eta_\beta)\hat\beta_t(z)
+\eta_\beta r_t(z), \\
\hat\beta_{t+1}(z)
&=
\tilde\beta_{t+1}(z)
-
\frac{1}{|\mathcal A|}
\sum_{a\in\mathcal A}\tilde\beta_{t+1}(a).
\tag{43}
\end{align}
We initialize $\hat\beta_0(z)=0$. The centering step imposes
$\sum_{z\in\mathcal A}\hat\beta_t(z)=0$ and separates the global arm effect from
the problem intercept. Importantly, Eqs.~(42)--(43) use only the $n_0$ probe
rollouts and never the adaptively allocated samples.

\paragraph{Coverage feature.}
After probing all arms, we form
\[
O^{\mathrm{probe}}_{q,t}
=
\bigcup_{a\in\mathcal A}\mathrm{probe}_t(q,a)
\]
and compute $\Omega_t(q,z)$ as in Eq.~(9). For guided arms, let
\[
\tilde g_t(q,z)
=
\log\!\left(\Omega_t(q,z)+\epsilon_\Omega\right).
\]
We remove its minibatch-wide offset,
\begin{equation}
\hat g_t(q,z)
=
\tilde g_t(q,z)
-
\frac{1}{|\mathcal B||\mathcal Z|}
\sum_{q'\in\mathcal B}
\sum_{z'\in\mathcal Z}
\tilde g_t(q',z'),
\qquad
\hat g_t(q,\varnothing)=0.
\tag{44}
\end{equation}
Centering makes $\omega_t$ capture whether relative differences in uncovered,
strategy-addressable mass predict arm success rather than absorbing the global
success-rate intercept.

\paragraph{Online coverage coefficient.}
We treat $\omega_t$ as a one-dimensional non-negative logistic coefficient. For a
candidate value $\omega$, define
\begin{equation}
m_{t,\omega}(q,z)
=
\sigma\!\left(
\hat c_t^{-z}(q)
+
\hat\beta_t(z)
+
\omega\hat g_t(q,z)
\right).
\tag{45}
\end{equation}
Using the current probe batch only to update the coefficient for the \emph{next}
training step, we perform projected stochastic gradient ascent on the regularized
binomial log-likelihood:
\begin{equation}
\omega_{t+1}
=
\Pi_{[0,\omega_{\max}]}
\left[
\omega_t
+
\eta_\omega
\left(
\frac{1}{|\mathcal B||\mathcal Z|}
\sum_{q\in\mathcal B}
\sum_{z\in\mathcal Z}
\hat g_t(q,z)
\bigl[
s_{q,z}-n_0m_{t,\omega_t}(q,z)
\bigr]
-
\lambda_\omega\omega_t
\right)
\right].
\tag{46}
\end{equation}
We initialize $\omega_0=0$. Hence, when the predicted coverage opportunity carries
little information about empirical success, its expected gradient vanishes and the
shrinkage term drives $\omega_t$ toward zero; the prior then reduces to the problem
and global-arm effects. At step $t$, both $\hat\beta_t$ and $\omega_t$ therefore
depend only on preceding probe batches, while $\hat c_t^{-z}(q)$ and
$\hat g_t(q,z)$ summarize the current all-arm probe.

\subsection{Discovered Strategy Prompts}
\label{app:strategy_prompts}

The following five natural-language instructions are used to instantiate the
strategy-conditioned arms identified by our discovery and certification
procedure. Each prompt specifies a reusable high-level reasoning procedure
rather than providing problem-specific solution information.

\paragraph{Stepwise Decomposition.}
\begin{quote}
\textit{
Break the problem into explicit sequential steps or subgoals. Solve each step
in order, verifying intermediate results by substitution or logical checks
before proceeding. Combine the results to obtain the final answer.
}
\end{quote}

\paragraph{Equation Setup.}
\begin{quote}
\textit{
Define variables for unknowns, translate the problem into equations or
formulas, then solve step-by-step, verifying the solution against the original
conditions.
}
\end{quote}

\paragraph{Formula Application.}
\begin{quote}
\textit{
Identify the relevant formula or theorem, state it explicitly, substitute the
given values, and compute step-by-step to obtain the final answer.
}
\end{quote}

\paragraph{Backward Reasoning.}
\begin{quote}
\textit{
Start from what the final answer must satisfy, work backwards to determine
what the earlier quantities must be, then verify the chain forward.
}
\end{quote}

\paragraph{Counterfactual Reasoning.}
\begin{quote}
\textit{
Consider a nearby alternative to one of the problem conditions and reason
through how the solution would change. Use this contrast to identify the
quantities, constraints, or relationships that are essential, then return to
the original conditions and solve the problem accordingly.
}
\end{quote}

\subsection{Fixed Strategy Baseline}
\label{app:fixed_strategy}

To disentangle the benefit of our discovered strategy repertoire from the generic effect of conditioning rollouts on diverse reasoning instructions, we construct a \textit{Fixed Strategy} baseline using an externally specified, model-agnostic strategy set. Each strategy is represented only by a short natural-language instruction. We do not use demonstrations, few-shot examples, solution prefixes, problem-specific hints, or privileged solution information. The same five instructions are fixed across all problems, model scales, datasets, and training runs.

\paragraph{Intuitive reasoning.}
\begin{quote} \textit{
Approach the problem by first building an intuitive understanding of the quantities, relationships, or structure involved. Use simple mental models and qualitative expectations to identify a plausible route to the solution, then turn that intuition into the necessary calculations or deductions.}
\end{quote}

\paragraph{Planning.}
\begin{quote} \textit{
Before carrying out detailed calculations, formulate a high-level plan for solving the problem. Identify the main intermediate goals and the order in which they should be addressed, then execute the plan systematically and revise it if necessary.}
\end{quote}

\paragraph{Analogical reasoning.}
\begin{quote} \textit{
Look for a simpler or more familiar problem with the same underlying structure. Reason through the analogous situation, identify which relationships carry over, and map those insights back to the original problem to obtain the solution.}
\end{quote}

\paragraph{Socratic reasoning.}
\begin{quote} \textit{
Guide the solution through a sequence of focused self-questions. At each stage, ask what is known, what must be determined next, and what follows from the current information; answer each question before proceeding to the next deduction.}
\end{quote}

\paragraph{Contrastive reasoning.}
\begin{quote} \textit{
Consider multiple plausible solution paths, interpretations, or candidate answers rather than committing immediately to one. Compare them using the constraints of the problem, eliminate inconsistent alternatives, and continue with the candidate best supported by the evidence.}
\end{quote}

For all five strategies, we use the same strategy-conditioning template:
\begin{quote}
\ttfamily
\{question\}

\vspace{0.5em}
Use the following approach as your primary reasoning strategy:

\vspace{0.5em}
\{strategy\_text\}

\vspace{0.5em}
Apply the approach naturally; do not mention this instruction.\\
Put your final answer in \textbackslash boxed\{\}.
\end{quote}

\subsection{Training and Implementation Details}
\label{app:training_details}

\paragraph{Models and parameterization.}
We evaluate PSRA with Qwen2.5-Instruct models at three scales:
1.5B, 3B, and 7B parameters. All models are trained using LoRA with rank
$32$, scaling factor $\alpha=64$, and adapters applied to all linear layers.
Model computation uses bfloat16 precision, SDPA attention, and gradient
checkpointing. We use AdamW with learning rate $3\times10^{-5}$,
gradient-norm clipping at $1.0$, and random seed $42$. Unless otherwise
specified, training runs for at most $1000$ optimization steps.

The model- and dataset-specific configurations are summarized below:
\begin{table}[h]
\centering
\small
\setlength{\tabcolsep}{4pt}
\begin{tabular}{lcccc}
\toprule
& 7B / DeepScaleR
& 7B / DAPO
& 3B / DAPO
& 1.5B / DAPO \\
\midrule
Backbone
& Qwen2.5-7B
& Qwen2.5-7B
& Qwen2.5-3B
& Qwen2.5-1.5B \\

Training data
& DeepScaleR-40K
& DAPO-Math-17K
& DAPO-Math-17K
& DAPO-Math-17K \\

Training split
& 7B pass@8 mixed
& 7B pass@8 mixed
& 3B pass@8 mixed
& 1.5B pass@8 mixed \\

GPUs
& $4\times$ GH200
& $4\times$ GH200
& $4\times$ GH200
& $4\times$ GH200 \\

Prompts / rank / step
& 32 & 8 & 8 & 8 \\

Global prompts / step
& 128 & 32 & 32 & 32 \\

LoRA rank / $\alpha$
& 32 / 64
& 32 / 64
& 32 / 64
& 32 / 64 \\

Learning rate
& $3\times10^{-5}$
& $3\times10^{-5}$
& $3\times10^{-5}$
& $3\times10^{-5}$ \\

Maximum steps
& 1000 & 1000 & 1000 & 1000 \\
\bottomrule
\end{tabular}
\caption{Training configurations across model scales and training corpora. All PSRA-specific hyperparameters and per-prompt rollout budgets are shared across settings.}
\label{tab:training_config}
\end{table}

\paragraph{Problem--strategy rollout allocation.}
The final strategy repertoire contains five certified strategies. Together
with the unguided condition, PSRA therefore operates over
\[
|\mathcal Z|=5,
\qquad
|\mathcal A|=|\mathcal Z|+1=6
\]
arms for each problem. Every active arm first receives $n_0=2$ probe
rollouts, resulting in a fixed probe cost of
\[
|\mathcal A|n_0=12
\]
rollouts per prompt. We use prior strength $\kappa=4$ and update the global
arm effect using an EMA coefficient $\eta_\beta=0.05$.

After probing, PSRA allocates an additional per-prompt-equivalent budget
$b_{\mathrm{extra}}=10$ across problem--arm pairs, subject to a maximum of
$n_{\max}=8$ rollouts per arm. For minibatch $\mathcal B$,
\[
\sum_{q\in\mathcal B}\sum_{z\in\mathcal A} k_{q,z}
\le
|\mathcal B|b_{\mathrm{extra}}.
\]
The total generation budget is therefore equivalent to
\[
|\mathcal A|n_0+b_{\mathrm{extra}}
=
6\times2+10
=
22
\]
policy rollouts per training prompt on average. The additional budget is
shared across $(q,z)$ pairs rather than assigned independently to each prompt.

\paragraph{Empirical-Bayes allocation prior.}
For each problem--arm pair, the prior mean is
\[
m_t(q,z)
=
\sigma\!\left(
\hat c_t^{-z}(q)
+
\hat\beta_t(z)
+
\omega_t\hat g_t(q,z)
\right),
\]
where $\hat c_t^{-z}(q)$ is the leave-one-arm-out problem effect,
$\hat\beta_t(z)$ is the centered probe-only global arm effect, and
$\hat g_t(q,z)$ is the centered log coverage feature. The estimators and the
online update of the non-negative coverage coefficient $\omega_t$ are given
in Appendix~\ref{app:online_prior}. Adaptively allocated rollouts are not used to update these prior statistics.

\paragraph{Policy optimization.}
Within each $(q,z)$ arm, advantages are normalized using rollouts from that
arm. PPO clipping uses $\epsilon=0.2$. For strategy-conditioned rollouts, the
off-context importance weight is clipped symmetrically in log space with
$w_{\max}=2$, corresponding to
\[
w(y)\in[1/2,2].
\]
The main experiments use a fixed preservation coefficient
\[
\lambda_t=\lambda_0=0.3,
\]
and adaptive preservation is disabled.

All model scales use the same KL regularization toward the frozen base policy,
with coefficient $\beta_{\mathrm{KL}}=0.04$. For LoRA training, the reference
policy is obtained by disabling the LoRA adapters, and KL is estimated using
the $k_3$ estimator.

\paragraph{Sampling and sequence limits.}
All training rollouts are sampled with temperature $1.0$ and top-$p=1.0$.
We use a maximum prompt length of $1024$ tokens and a maximum completion
length of $2048$, giving a maximum sequence length of $3072$. No explicit
length shaping is used ($\alpha_{\mathrm{len}}=0$). Responses terminated by
the maximum completion length receive a truncation penalty of $0.2$ in the
policy-training score. The Bayesian allocator itself uses binary answer
correctness when computing the success counts $s_{q,z}$.

\paragraph{Distributed training and rollout budgets.}
All experiments use four GH200 GPUs with data parallelism, tensor-parallel
size $1$, and colocated vLLM rollout generation. Each rank independently
samples its local prompt minibatch, while gradients are averaged across all
four ranks.

For the 7B DeepScaleR-40K setting, we use $32$ prompts per rank, giving a
global minibatch of $128$ prompts. Under the average budget of $22$ rollouts
per prompt, this corresponds to
\[
128\times22=2816
\]
rollout slots per optimization step, or $704$ per rank.

For the 7B DAPO-Math-17K setting and the 3B and 1.5B scale-study runs, we use
a split-batch configuration with $8$ prompts per rank and $32$ prompts
globally. The corresponding generation budget is
\[
32\times22=704
\]
rollout slots per optimization step, or $176$ per rank. Thus, all settings
share the same per-prompt exploration budget and PSRA hyperparameters, while
the DeepScaleR 7B run uses a larger global prompt batch.

The sufficient statistics used for the global arm effect and online coverage
coefficient are synchronized across data-parallel workers before updating the
shared allocator state.

\paragraph{Baseline reproduction and compute matching.}
For Vanilla GRPO, we use the same backbone, training data, LoRA
parameterization, optimizer, learning rate, sampling configuration, sequence
limits, KL regularization, global prompt batch, and number of optimization
steps as the corresponding PSRA run. The only differences are therefore the
PSRA-specific strategy conditioning, rollout allocation, transfer correction,
and preservation objective. In particular, GRPO is matched to the same total
generation budget, equivalent to $22$ policy rollouts per training prompt on
average.

For the remaining baselines, we reproduce each method using the
method-specific hyperparameters and optimization settings reported in its
public implementation or paper whenever available. To ensure a compute-matched
comparison, we adjust the total rollout budget and the number of optimizer
updates to match the corresponding PSRA setting, while keeping the backbone,
training corpus, evaluation protocol, and inference budget identical.
Because guided-RL baselines can be sensitive to implementation-specific
hyperparameters and exhibit substantial run-to-run training variation, we
perform three tuning runs for each reproduced baseline and report the
configuration selected by the best held-out validation performance.

\paragraph{Evaluation setup.}
We evaluate all models with the same decoding and answer-verification pipeline. Each problem is formatted as a user-only chat message with the instruction \texttt{Put your final answer in \textbackslash boxed\{\}}, and generation uses temperature $0.6$ and top-$p=0.95$. Mathematical answers are extracted only from the final \texttt{\textbackslash boxed\{\}} expression and verified with
Math-Verify against normalized gold LaTeX answers; no bare-expression fallback is used. All evaluations use deterministic question- and sample-specific random seeds.

For the main evaluation, we draw $16$ independent responses per problem and
estimate Pass@1 by the mean single-sample correctness. Pass@$k$ is computed
using the standard unbiased estimator. 
Generation is capped at $4096$ new tokens for MATH-500, AMC 2023,
Minerva Math, StrategyQA, GPQA-Diamond, and MMLU-Pro, and at $8192$ tokens
for AIME and Omni-MATH. For the inference-budget analysis, we additionally sample $128$ responses per problem on AIME 2024/2025, AMC 2023, and MATH-500, and report $k\in\{1,2,4,8,16,32,64,128\}$. All methods use
identical prompts, decoding parameters, evaluation subsets, generation limits,
and scoring code.

\subsection{Qualitative Examples of Strategy Switching}
\label{app:strategy_examples}

We provide two representative examples in which the default reasoning route fails while an alternative strategy yields a correct solution. For each problem, we first show the original rollouts and then summarize their strategy-level reasoning structure using the five strategy families identified in Section~\ref{sec:strategy_analysis}.

\subsubsection{Example 1: Formula Application to Stepwise Decomposition}

\paragraph{Problem.}
Paul and Jesse each choose a number at random from the first six primes. What is the probability that the sum of the numbers they choose is even?

\paragraph{Failed rollout: \texttt{formula\_application}.}
\begin{quote}\small
To determine the probability that the sum of the numbers chosen by Paul and Jesse is even, we first need to identify the first six prime numbers. These are: $2,3,5,7,11,$ and $13$.

The sum of two numbers is even if both numbers are either even or both are odd. Among the first six prime numbers, only $2$ is even, and the rest ($3,5,7,11,13$) are odd. Therefore, the only way to get an even sum is if both Paul and Jesse choose the number $2$.
The probability that Paul chooses $2$ is $\frac{1}{6}$. The probability that Jesse chooses $2$ is $\frac{1}{6}$.

Since the choices are independent, the probability that both Paul and Jesse choose $2$ is
\[
\frac{1}{6}\times\frac{1}{6}=\frac{1}{36}.
\]

Thus, the probability that the sum of the numbers they choose is even is
\[
\boxed{\frac{1}{36}}.
\]
\end{quote}

\paragraph{Successful alternative rollout: \texttt{stepwise\_decomposition}.}
\begin{quote}\small
To determine the probability that the sum of the numbers chosen by Paul and Jesse is even, we first need to identify the first six prime numbers. These are: $2,3,5,7,11,$ and $13$.

The sum of two numbers is even if both numbers are either even or both are odd. Among the first six prime numbers, only one number is even ($2$), and the rest are odd ($3,5,7,11,13$).

Let's calculate the total number of possible outcomes. Since each person chooses a number independently from the six primes, there are
\[
6\times6=36
\]
possible outcomes.

Next, we count the favorable outcomes where the sum is even. This can happen in two scenarios:
\begin{enumerate}
    \item Both numbers chosen are even.
    \item Both numbers chosen are odd.
\end{enumerate}

For the first scenario, the only even number is $2$. Hence there is only one favorable outcome.

For the second scenario, Paul and Jesse can each choose from the five odd numbers, giving
\[
5\times5=25
\]
favorable outcomes.

Thus, the total number of favorable outcomes is
\[
1+25=26,
\]
and the probability is
\[
\frac{26}{36}=\frac{13}{18}.
\]

Therefore,
\[
\boxed{\frac{13}{18}}.
\]
\end{quote}

\paragraph{Strategy-level structure.}
\[
\begin{array}{ll}
\textbf{Failed:} &
\text{identify parity rule}
\rightarrow
\text{focus on the unique even prime}
\rightarrow
P(2,2)
\rightarrow
\frac{1}{36},
\\[4pt]
\textbf{Successful:} &
\text{decompose the target event}
\rightarrow
\{\text{even--even},\,\text{odd--odd}\}
\rightarrow
1+25
\rightarrow
\frac{13}{18}.
\end{array}
\]

The failed route applies the parity condition directly but prematurely collapses the event to the even--even case. The alternative strategy explicitly decomposes the target event into exhaustive subcases before combining their contributions.

\subsubsection{Example 2: Equation Setup to Formula Application}

\paragraph{Problem.}
In triangle $ABC$, $AB=AC=5$ and $BC=6$. Let $O$ be the circumcenter of triangle $ABC$. Find the area of triangle $OBC$. The answer is in the form $\frac{m}{n}$, where $\gcd(m,n)=1$; report $m+n$.

\paragraph{Failed rollout: \texttt{equation\_setup}.}
\begin{quote}\small
To find the area of triangle $OBC$, we start by determining the coordinates of the vertices of triangle $ABC$ and the circumcenter $O$.

First, place $B$ at $(-3,0)$ and $C$ at $(3,0)$. Since $AB=AC=5$, the coordinates of $A$ are $(0,4)$.

Next, we find the circumcenter $O$. The circumcenter is the intersection of the perpendicular bisectors of the sides of the triangle. The perpendicular bisector of $BC$ is the vertical line
\[
x=0.
\]

The midpoint of $AB$ is $(-1.5,2)$ and the slope of $AB$ is $-\frac{4}{3}$, so the slope of its perpendicular bisector is $\frac{3}{4}$. Its equation is
\[
y-2=\frac34(x+1.5)
\]
and hence
\[
y=\frac34x+\frac{15}{8}+2
 =\frac34x+\frac{23}{8}.
\]

Setting $x=0$, we obtain
\[
y=\frac{23}{8},
\]
so
\[
O=\left(0,\frac{23}{8}\right).
\]

The base $BC$ is $6$, and the height from $O$ to $BC$ is $\frac{23}{8}$. Therefore,
\[
[OBC]
=
\frac12\cdot6\cdot\frac{23}{8}
=
\frac{69}{8}.
\]

Thus,
\[
m+n=69+8=\boxed{77}.
\]
\end{quote}

\paragraph{Successful alternative rollout: \texttt{formula\_application}.}
\begin{quote}\small
To find the area of triangle $OBC$, we start by determining the circumradius $R$ of triangle $ABC$. Since $AB=AC=5$ and $BC=6$, triangle $ABC$ is isosceles.

Let $D$ be the midpoint of $BC$. Then
\[
BD=DC=3.
\]
In right triangle $ABD$,
\[
AD^2+BD^2=AB^2,
\]
so
\[
AD^2+3^2=5^2
\quad\Longrightarrow\quad
AD=4.
\]

Therefore, the area of triangle $ABC$ is
\[
K=\frac12\cdot6\cdot4=12.
\]

Using the circumradius formula
\[
R=\frac{abc}{4K},
\]
we obtain
\[
R
=
\frac{6\cdot5\cdot5}{4\cdot12}
=
\frac{25}{8}.
\]

Since $O$ is the circumcenter, $OB=R$. In right triangle $OBD$,
\[
OB^2=OD^2+BD^2,
\]
so
\[
\left(\frac{25}{8}\right)^2
=
OD^2+3^2.
\]
Hence
\[
OD^2
=
\frac{625}{64}-\frac{576}{64}
=
\frac{49}{64},
\qquad
OD=\frac78.
\]

Finally,
\[
[OBC]
=
\frac12\cdot BC\cdot OD
=
\frac12\cdot6\cdot\frac78
=
\frac{21}{8}.
\]

Thus,
\[
m+n=21+8=\boxed{29}.
\]
\end{quote}

\paragraph{Strategy-level structure.}
\[
\begin{array}{ll}
\textbf{Failed:} &
\text{choose coordinates}
\rightarrow
\text{construct perpendicular bisectors}
\rightarrow
\text{solve for }O
\rightarrow
[OBC],
\\[4pt]
\textbf{Successful:} &
AD
\rightarrow
[ABC]
\rightarrow
R=\frac{abc}{4K}
\rightarrow
OD=\sqrt{R^2-BD^2}
\rightarrow
[OBC].
\end{array}
\]

Here the unsuccessful route translates the geometry into a coordinate system and explicitly solves for the circumcenter, whereas the alternative route avoids recovering $O$ altogether and instead composes standard geometric formulas for the circumradius and the distance from the circumcenter to the chord.

Together, the two examples illustrate that a failed rollout need not indicate the absence of a viable solution procedure. Switching the strategy changes how the same problem is organized---from direct formula use to explicit case decomposition in the first example, and from coordinate-based equation setup to theorem-driven formula application in the second---exposing successful trajectories that remain accessible to the same model.

\end{document}